\documentclass[runningheads]{llncs}
\usepackage[T1]{fontenc}
\usepackage{graphicx}
\usepackage{booktabs}
\usepackage{hyperref}
\usepackage[numbers]{natbib}
\usepackage[misc]{ifsym}
\usepackage{float}
\newcommand{\corr}{(\Letter)}
\begin{document}

\title{Little Brains, Big Feats: Exploring Compact Language Models}

\titlerunning{Little Brains, Big Feats: Exploring Compact Language Models}

\author{Dari Baturova \and
Elena Bruches
 \and
Ivan Chernov \and
Roman Derunets \corr \and
Arsenii Fomin \and
Andrey Kostin
}
\institute{Siberian Neuronets LLC, Novosibirsk, Russia \\
\email{r.derunets@alumni.nsu.ru}}

\authorrunning{Baturova et al.}

\tocauthor{Dari Baturova, Elena Bruches, Ivan Chernov, Roman Derunets, Arsenii Fomin, and Andrey Kostin}
\toctitle {Little Brains, Big Feats: Exploring Compact Language Models}

\maketitle

\begin{abstract}
While large language models have been dominating the research landscape recently, small language models remain highly relevant across various domains; yet, they receive far less attention. In this study, we investigate how smaller language models perform during the generation stage within a Retrieval-Augmented Generation (RAG) system. To benchmark these models effectively, we utilised both open-source and proprietary datasets covering diverse subject areas and question types.
Our findings demonstrate that a RAG system with small language models can be executed directly on-device without requiring any GPU hardware within a reasonable time. The experimental code and links to the supplementary materials can be accessed through the GitHub repository\footnote{\url{https://github.com/SibNN/SLM-RAG-EVAL}}.

\keywords{Small Language Models \and Benchmarking \and Retrieval-Augmented Generation}
\end{abstract}

\section{Introduction}

In recent years, Large Language Models (LLMs) have demonstrated remarkable performance across various tasks. However, certain limitations persist, including their inability to rapidly process novel data and effectively manage domain-specific facts. To address these challenges, the RAG approach has been introduced \cite{10.5555/3495724.3496517}. Conventionally, a RAG system comprises two primary components: Retrieval and Generation.

The Retrieval module focuses on identifying the most relevant segments within external knowledge storage --- such as databases --- in accordance with user queries. Meanwhile, the Generation component leverages retrieved information alongside the user's input to generate accurate responses.

Although Retrieval utilises smaller-scale language models to compute text embeddings for both database content and inputs, the Generation phase relies heavily on LLMs to deliver high-quality output. Furthermore, embedding models require significantly fewer computational resources since they encode the knowledge base once. Conversely, LLMs employed during the Generation step demand substantial computing power. Consequently, when sufficient resources or budgetary allocations are unavailable, employing Small Language Models (SLMs) could provide a viable alternative solution.

Small Language Models (SLMs) are compact AI models that contain significantly fewer parameters -- ranging from millions to several billions -- compared to LLMs. These smaller models are engineered to operate efficiently in resource-constrained environments like edge devices, smartphones, or personal computers, requiring less computational power, memory storage, and energy consumption. Their advantages include faster processing times, enhanced privacy due to local operation (often functioning offline), and optimisation for specialised tasks within particular domains instead of broad general-purpose reasoning capabilities.

These models proved to be beneficial in scenarios where computational resources are insufficient to support larger-scale LLMs, particularly on edge computing platforms \cite{10.1145/3724420}. Additionally, SLMs become essential when handling sensitive information that cannot be transmitted through APIs to external service providers \cite{10.1145/3768165}. Such situations necessitate deploying the model directly onto internal hardware, which frequently lacks sufficient capacity to execute highly parameterised models effectively. For instance, executing a model with as many as 14 billion parameters demands substantial computational resources typically beyond the reach of standard end-users. By comparison, SLMs demand far fewer resources, enabling deployment even on CPUs, thereby expanding their applicability across diverse settings.

\begin{figure}[h]
\centering
\includegraphics[width=1.0\linewidth]{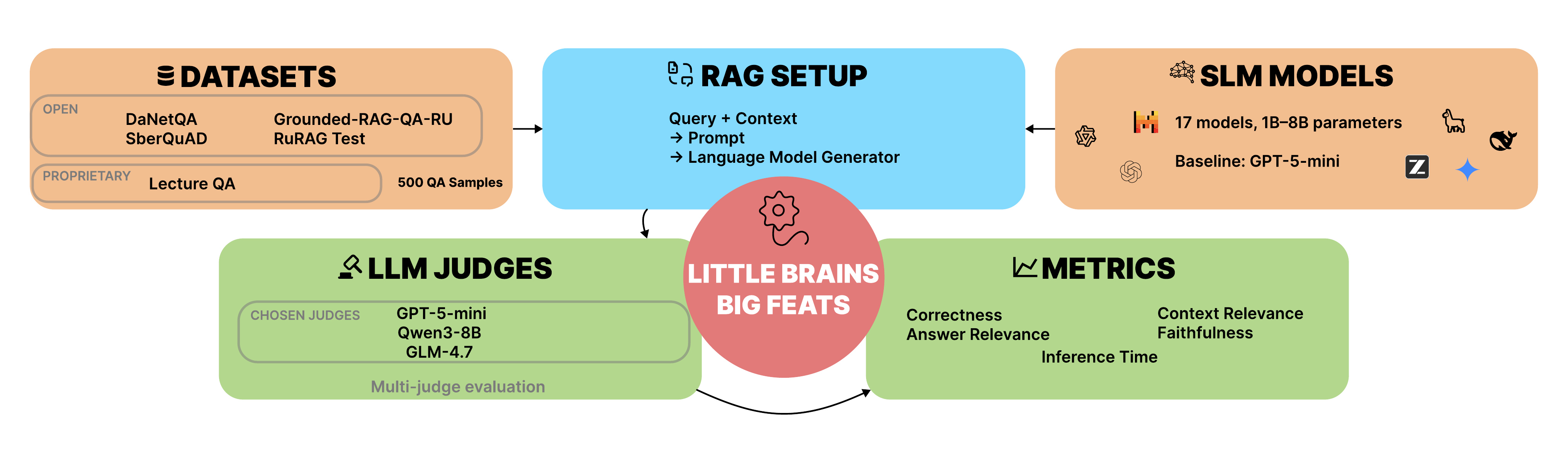}
\caption{Overview of the evaluation pipeline. The benchmark combines five Russian-language QA datasets. Small language models generate answers in a RAG setting, and responses are evaluated using a multi-judge LLM-as-a-Judge framework across several quality metrics.}
\label{fig:overview}
\end{figure}

In this work, we investigate the generation capabilities of SLMs within our product-oriented RAG framework, which features local CPU language model inference. To enable a comprehensive evaluation, we construct a benchmark consisting of both open-source and proprietary datasets that cover diverse subject areas and question types. Our results demonstrate that RAG systems built with SLMs can operate efficiently without GPU hardware, making fully on-device deployment feasible for many applications. The overall system is illustrated in Figure~\ref{fig:overview}.

The main contributions of this work are summarised as follows:
\begin{itemize}
    \item \textbf{Dataset construction:} We assemble a Russian-language benchmark that combines available open-source and proprietary sources to evaluate retrieval-augmented generation performance.
    \item \textbf{Model benchmarking:} We conduct a systematic evaluation of SLMs within a RAG framework.
    \item \textbf{Extensive analysis:} We provide a detailed analysis of performance characteristics, evaluation methods, and practical considerations for deploying SLMs in on-device RAG systems.
\end{itemize}

\section{Related Work}

\subsection{Retrieval-Augmented Generation}

Retrieval-Augmented Generation (RAG) combines information retrieval techniques with neural text generation to improve factual accuracy and access to external knowledge. Early open-domain question answering systems adopted a retrieve-and-read paradigm, where relevant documents were first retrieved and then processed by a neural reader model to extract answers \cite{chen2017reading}. 

Traditional retrieval methods relied on lexical matching approaches such as TF-IDF \cite{ramos2003using} and BM25 \cite{robertson2009probabilistic}. While computationally efficient, these techniques are limited in their ability to capture semantic similarity between queries and documents. To address this limitation, neural retrieval models based on dense representations were introduced. Dense Passage Retrieval (DPR) \cite{karpukhin2020dense} demonstrated that dual-encoder architectures trained with contrastive objectives significantly outperform classical sparse retrieval methods on open-domain QA tasks. Other approaches, such as ColBERT \cite{khattab2020colbert}, further improved retrieval quality by enabling late interaction between contextual token embeddings.

Building on these developments, retrieval-augmented models integrate retrieval directly into the generation process. The RAG framework proposed by Lewis et al. \cite{10.5555/3495724.3496517} combines a neural retriever with a sequence-to-sequence generator, allowing the model to condition its outputs on external documents retrieved at inference time. Similar ideas were explored in REALM \cite{guu2020retrieval}, which incorporates retrieval during language model pretraining to enable models to access large external knowledge sources.

Recent work has also extended RAG beyond purely textual question answering. For example, multimodal retrieval-augmented generation can be used to convert a text-based language model into a multimodal system by retrieving external knowledge from text, image, and audio inputs, without requiring additional resource-intensive multimodal training~\cite{derunets546knowledge}. This line of work demonstrates that retrieval can expand model capabilities in complex settings while reducing the need for heavyweight end-to-end training.

At the same time, other studies focus on improving the generation stage through the use of large language models and their ensembles~\cite{bondarenko2026raguteamsemeval2026task8}. 
Our study is complementary to these directions: instead of increasing the complexity of the generator, the input modalities, or the overall pipeline, we investigate whether small language models can serve as effective generators in a retrieval-augmented setting under limited computational resources.

\subsection{Small Language Models}

Small Language Models (SLMs) are compact neural language models (hundreds of millions to a few billion parameters) designed for low-latency, low-memory inference on CPUs, mobile devices, and edge devices; they trade scale for efficiency while preserving practical NLP performance.

Compression and knowledge transfer remain the dominant training patterns: knowledge-distillation and task-aware distillation have produced widely used SLMs such as DistilBERT \cite{sanh2019distilbert} and MobileBERT \cite{sun2020mobilebertcompacttaskagnosticbert}, which demonstrate large teacher-small student gains in latency and size.

More recent families show end-to-end SLM pretraining and strong-to-weak distillation (examples: TinyLlama \cite{zhang2024tinyllamaopensourcesmalllanguage}, industry-driven smaller variants in the Qwen \cite{yang2025qwen3technicalreport} family, and purpose-optimised models such as Shakti \cite{shakhadri2025shakti25billionparameter}). These models target on-device assistants, privacy-preserving analytics, and industrial automation.

In science and industry, SLMs are widely used for domain-specific chatbots, on-device document processing, real-time IoT and predictive-maintenance analytics, and privacy-preserving biomedical or legal question answering systems \cite{kandala2024tinyllm,10.1145/3768165}. Such deployments typically combine model compression techniques such as quantisation \cite{jin-etal-2024-comprehensive}, parameter-efficient fine-tuning methods including LoRA \cite{lora} and adapter-based approaches, synthetic data generated by larger teacher models, and inference optimisations designed to meet strict latency and energy constraints \cite{peft_survey}. Frameworks and empirical studies focused on edge-oriented language models, such as TinyLLM, provide practical guidelines and evaluation pipelines for training and deploying SLMs in real-world production environments \cite{kandala2024tinyllm}.

\subsection{Benchmarks}

There exists a variety of datasets tailored for evaluating Retrieval-Augmented Generation (RAG) systems. Among these, one of the most extensively utilised benchmarks is RAGBench \cite{friel2025ragbenchexplainablebenchmarkretrievalaugmented}, comprising 100,000 examples spread across five distinct industrial domains and encompassing diverse RAG task types. Another prominent framework is CRAG \cite{yang2024cragcomprehensiverag}, which serves as a factual question-answering benchmark featuring 4,409 question-answer pairs along with mock APIs for web searches and knowledge graph retrievals.

Moreover, several specialised datasets have been proposed for evaluating specific RAG scenarios, including LegalBench-RAG \cite{pipitone2024legalbenchragbenchmarkretrievalaugmentedgeneration}, which focuses on legal retrieval-augmented generation; REAL-MM-RAG \cite{wasserman-etal-2025-real}, which addresses multimodal RAG challenges; DRAGOn \cite{chernogorskii2026dragondesigningragperiodically}, which targets periodically updated RAG settings; MultiHop-RAG \cite{tang2024multihopragbenchmarkingretrievalaugmentedgeneration}, which evaluates complex multi-hop reasoning tasks; and MTRAG \cite{katsis2025mtragmultiturnconversationalbenchmark}, which is designed for multi-turn conversational benchmarking.

However, Small Language Models (SLMs) possess unique characteristics (such as domain specificity and the ability to run under limited computational resources) that distinguish them from their larger counterparts, Large Language Models (LLMs). As such, their evaluation process should diverge accordingly. While traditional benchmarking procedures remain applicable, efforts have been made to create specialised frameworks more aligned with SLMs' inherent properties.

One illustrative example is SLM-Bench \cite{pham-etal-2025-slm}, a benchmark explicitly designed to evaluate SLMs across multiple dimensions, including accuracy, computational efficiency, and environmental sustainability. Comprising nine Natural Language Processing (NLP) tasks -- such as classification, question answering, and sentiment analysis -- and utilising 23 datasets spanning 14 diverse domains (from common sense to physics, video gaming, and news), this benchmark offers a holistic assessment of SLMs' strengths and limitations. Similarly, HealthSLM-Bench \cite{wang2025healthslmbenchbenchmarkingsmalllanguage} focuses on health prediction tasks across three real-world mobile and wearable datasets, showcasing SLMs' utility in biomedical contexts. SLMQuant \cite{10.1145/3746262.3761973} introduces a systematic methodology for assessing compression techniques applied to SLMs, employing rigorous multi-dimensional evaluations across varied architectures and tasks to analyse state-of-the-art quantisation methods.

In summary, contemporary trends in benchmark development emphasise coverage across a broad spectrum of domains and tasks. However, to the best of our knowledge, there remains a significant gap regarding Russian-language RAG benchmarks capable of comprehensively covering diverse linguistic nuances and application scenarios.

\subsection{LLM-as-a-Judge Evaluation}

Evaluating generative systems such as LLM-based question answering or RAG pipelines remains a challenging task. Traditional automatic metrics including BLEU \cite{papineni2002bleu}, ROUGE \cite{lin2004rouge}, and METEOR \cite{banerjee2005meteor} rely on lexical overlap between generated outputs and reference texts. While these metrics are effective for machine translation or summarisation, they often fail to capture semantic correctness, factual grounding, and reasoning quality in open-ended generation tasks.

To address these limitations, recent work explores the use of large language models themselves as automated evaluators, a paradigm commonly referred to as \textit{LLM-as-a-Judge}. One of the most influential approaches is G-Eval \cite{liu2023g}, which leverages a strong LLM (e.g., GPT-4) to assess generated responses according to predefined evaluation criteria such as coherence, factuality, and relevance. The method formulates evaluation as a structured reasoning process where the model first produces intermediate evaluation steps before assigning a final score. Experimental results demonstrate a strong correlation between G-Eval scores and human judgments.

Similarly, frameworks such as MT-Bench and Chatbot Arena \cite{zheng2023judging} evaluate conversational systems through pairwise comparisons judged by LLMs. In this setup, the evaluator model compares responses from different systems and determines which one better satisfies the user query. This pairwise ranking approach reduces bias associated with absolute scoring and has been widely used for benchmarking modern chat models.

Several studies also explore specialised evaluation pipelines for RAG systems. For instance, RAGAS \cite{es2024ragas} introduces a reference-free evaluation framework that uses LLMs to assess dimensions such as answer relevance, faithfulness to retrieved documents, and context precision. By leveraging LLM reasoning capabilities, these approaches can detect hallucinations and grounding errors that traditional string-based metrics fail to capture.

Overall, LLM-as-a-Judge methodologies provide a scalable alternative to human evaluation and enable more nuanced assessment of generative systems, particularly for complex tasks such as retrieval-augmented question answering and conversational reasoning. However, challenges remain regarding evaluator bias, reproducibility, and sensitivity to prompt design, which continue to be active research directions.

\section{Data Description}

To evaluate the performance of small language models in the generation stage of a RAG pipeline, we constructed a benchmark consisting of multiple Russian-language question answering datasets. The benchmark includes both publicly available datasets and one proprietary dataset in order to cover a diverse range of domains and question types.

\subsection{Open-Source Datasets}

Below we briefly describe the publicly available datasets used in the benchmark.

\textbf{DaNetQA.}
 DaNetQA \cite{Glushkova_2021} is a Russian dataset consisting of yes/no questions paired with supporting text passages. Each example is represented as a triplet containing a question, a text fragment that potentially contains the answer, and a binary label indicating whether the statement is true or false. The dataset focuses on natural language inference, commonsense reasoning, and world knowledge.

\textbf{SberQuAD.}
SberQuAD \cite{Efimov_2020} is a Russian reading comprehension dataset. The dataset contains questions written by crowdworkers based on Wikipedia articles. Each question is associated with a passage where the answer appears as a text span, although some questions may be unanswerable.

\textbf{RuRAG Test Dataset.}
The RuRAG Test Dataset \cite{ruragdataset} was specifically designed for evaluating Russian-language RAG systems. It consists of questions, ground-truth answers, and contextual passages extracted from Russian Wikipedia articles.

\textbf{Grounded-RAG-QA-RU.}
Grounded-RAG-QA-RU \cite{groundedru} is a dataset designed to evaluate the ability of language models to answer questions using information grounded in provided documents. The dataset was generated using clusters of Russian Wikipedia articles and synthetic question-answer pairs produced with GPT-4. Questions may require reasoning across multiple documents, and some examples intentionally contain out-of-distribution queries that cannot be answered using the provided context. This setup encourages models to rely strictly on retrieved documents.

\subsection{Proprietary Dataset}

In addition to the open-source datasets described above, we included a proprietary dataset containing domain-specific question answering samples. Due to licensing restrictions, this dataset cannot be publicly released, but it follows the same structure as the other datasets, consisting of question, context, and reference answer fields.

The dataset consists of conference presentation texts and hand-crafted questions. The knowledge base contains 5,000 lecture presentations, each around 4,500 words long. It was created to evaluate the performance of a RAG system developed for an industrial application and deployed in production. Using these data, we evaluate the robustness of RAG systems in real-world scenarios where domain-specific knowledge is required.

\subsection{RAG Evaluation Dataset Construction}

In total, five datasets were used for evaluation. Before inclusion in the benchmark, each dataset underwent a preprocessing stage. First, all samples containing empty or incomplete fields were removed. Then, to ensure balanced evaluation across datasets, we randomly sampled 100 examples from each dataset, resulting in a final evaluation set of 500 samples.

To analyse model performance across different reasoning patterns, we further categorised the questions by type following \cite{10.1145/3477495.3531926}. Question classification was performed using the Qwen3-8B model, which assigned each question to one of several categories. The categories of questions include:

\begin{itemize}
\item \textbf{Factoid} --- questions requiring the retrieval of a specific factual piece of information;
\item \textbf{Reasoning} --- questions requiring logical inference or multi-step reasoning;
\item \textbf{Evidence-based} --- questions where the answer must be directly grounded in the provided context;
\item \textbf{Comparison} --- questions requiring a comparison between entities or concepts;
\item \textbf{Experience-based} --- questions involving subjective or experiential interpretation;
\item \textbf{Instruction} --- questions where the answer is supposed to be an instruction.
\end{itemize}

The distribution of these classes is shown in Table~\ref{tab:dataset_overview}.

\subsection{LLM-as-Judge Evaluation Dataset}
\label{sec:llm-as-judge-eval}
In addition to evaluating the RAG pipeline itself, we constructed a separate dataset to assess the reliability of an LLM-as-Judge evaluation approach. The goal of this dataset is to test whether an evaluator model can correctly distinguish between valid and invalid answers.

To create negative examples, we intentionally mismatched elements from different samples in the RAG evaluation dataset. For example, a question from one dataset is paired with an answer from a second dataset, a golden answer from a third, and context from a fourth. These artificially constructed combinations represent incorrect responses and were assigned a score of 0.

Positive examples were created by keeping the original pairs of context, question, and correct answer from the evaluation datasets. These samples represent valid responses and were assigned a score of 1.

Finally, positive and negative samples were combined into a single benchmark dataset. This mixed dataset exposes the evaluator model to both correct and clearly incorrect answers, enabling a more robust assessment of its ability to judge response quality.

\begin{table}[ht]
\centering
\caption{Overview of datasets used in the RAG evaluation benchmark. For each dataset, we report the number of sampled examples and distribution of question types.}
\begin{tabular}{lccccccc}
\hline
Dataset & Samples & Fact & Evidence & Reason & Comparison & Exp & Instruct  \\
\hline
DaNetQA & 100 & 79 & 2 & 16 & 1 & 2  & 0\\
Grounded-RAG-QA-RU & 100 & 36 & 18 & 4 & 25 & 17 & 0 \\
SberQuAD & 100 & 60 & 24 & 8 & 6 & 2 & 0 \\
ru-rag-test & 100 & 99 & 0 & 1 & 0 & 0 & 0 \\
Proprietary & 100 & 4 & 33 & 16 & 8 & 36 & 3 \\
\hline
Total & 500 & 278 & 77 & 45 & 40 & 57 & 3 \\
\hline
\end{tabular}
\label{tab:dataset_overview}
\end{table}

\subsection{Dataset Analysis}

As outlined earlier, our benchmark comprises 500 samples sourced from various datasets and distributed across different question types.

The mean question length amounts to 8.72 tokens (ranging from a minimum of 3 tokens to a maximum of 27 tokens), whereas the mean length for golden answers stands at 41.62 tokens (spanning from 1 token up to 364 tokens).

Additionally, we quantified the similarity among distinct data sources. Our findings revealed substantial diversity, indicating broad topic coverage and lexical variety within the final benchmark. Pairwise cosine similarities between these datasets ranged from 0.06 to 0.12, underscoring their heterogeneous nature.

Furthermore, an analysis of question complexity was conducted using the Qwen3-8B model. Questions were rated on a scale from 1 to 10. We obtained an overall average rating of 4.94, indicating that the questions are moderately challenging rather than overly simplistic.

Lastly, we evaluated the alignment between each question and its respective golden answer. Again employing the same LLM, we scored this correspondence on a scale of 1--10, achieving an average score of 7.012. This suggests high relevance and minimisation of potential errors in responses.

\section{Experiments}

\subsection{Models}

Candidate models were selected based on their parameter size, coverage of diverse model families, open-source availability, support for the GGUF format, and the ability to run within a 16 GB RAM constraint for local inference.

To ensure practical applicability, all candidate models were tested in a CPU-only environment. This preliminary evaluation allowed us to verify their compatibility with our system and confirm that inference could be executed correctly without GPU acceleration.

As a result, we identified 17 models that were suitable for further experiments. Additionally, GPT-5-mini was included as a state-of-the-art reference model for comparison.

\subsection{Generation Modes}

At the answer generation stage, we considered two modes:

\begin{itemize}
\item \textbf{Context mode}: The model receives the query together with the reference documents used as contextual information for answer generation.

\item \textbf{No-context mode}:  The model receives only the query without any additional context.
\end{itemize}

In this setup, we focus exclusively on the generation stage, assuming that the retrieval component has already produced the most relevant documents.

All candidate models were evaluated in the context mode. Additionally, one of the modern proprietary models (GPT-5-mini) was evaluated in both modes. This allows us to compare the performance of the candidate models against a strong baseline and to analyse the impact of contextual information on answer quality.

\subsection{Evaluation Setup}

We conducted the evaluation in two stages. First, we assessed multiple LLM-as-Judge models and selected the most reliable ones. Second, we evaluated language models used in the generation stage of the RAG pipeline.

The following LLM-as-Judge metrics were used to evaluate the RAG systems:
\begin{itemize}
    \item \textbf{Correctness} - whether the answer is factually correct.
    \item \textbf{Answer Relevance} - whether the answer addresses the user query.
    \item \textbf{Context Relevance} - whether the retrieved context is relevant to the user query.
    \item \textbf{Faithfulness} - whether the answer is supported by the provided context.
\end{itemize}
All judges provide scores in the continuous range $[0, 1]$ for each metric.

When evaluating candidate generation models, Context Relevance was excluded from the comparison because all models received the same retrieved documents for each query. Therefore, this metric was uninformative for model selection, but it was retained at the judge selection stage, where it served as one of the criteria for assessing judge reliability.

\subsection{Judge Evaluation}

To assess the performance of judges for RAG evaluation, we conducted a systematic benchmarking of multiple judges on the \textit{LLM-as-Judge Evaluation Dataset} as described in Section~\ref{sec:llm-as-judge-eval}. The main goal of this evaluation was to identify a subset of judges whose assessments are both reliable and well aligned with the overall consensus.

\subsubsection{Judge Performance Metrics}

For each judge and evaluation metric, we computed several metrics.

First, we measured the \textbf{F1 score} for distinguishing correct and incorrect answers. Continuous judge scores were binarised using a threshold of $0.5$.

Second, we computed the \textbf{Average Bad Score}, which reflects how strongly a judge penalises clearly incorrect answers. For judge $j$:

\[
AvgBadScore_{j} =
\frac{1}{|B|}\frac{1}{|M|} \sum_{i \in B}\sum_{m \in M} s_{i,j,m},
\]

where $B$ denotes the set of incorrect answers, $M$ denotes the set of evaluation metrics, and $s_{i,j,m}$ is the score assigned by judge $j$ to sample $i$ for metric $m$.

Lower values indicate that a judge assigns lower scores to incorrect responses and therefore better identifies errors.

Third, we measured the correlation between the judges on each metric using the Pearson correlation coefficient:

\[
\rho_{j, m} = \mathrm{corr}(s_{j,m}, \overline{s_m}),
\]

where $s_{j,m}$ denotes the score for judge $j$ on metric $m$, and $\overline{s_m}$ denotes the average score across all judges on the same examples for metric $m$.

\begin{table}[ht]
\centering
\caption{Judge evaluation results. The table reports the Average Bad Score (ABS), indicating how highly a judge scores incorrect responses; the F1 score for classifying correct vs.\ incorrect responses; and correlations between each judge metric and the consensus: Correctness correlation (C Corr), Answer Relevance correlation (AR Corr), Context Relevance correlation (CR Corr), and Faithfulness correlation (F Corr). Judges selected for final evaluation are underlined.}
\begin{tabular*}{\textwidth}{l@{\extracolsep{\fill}}cccccc}
\hline
Model Name & F1$\uparrow$ & C Corr$\uparrow$ & AR Corr$\uparrow$ & CR Corr$\uparrow$ & F Corr$\uparrow$ & ABS$\downarrow$ \\
\hline
DeepSeek-v3.2-alt & 1.00 & 1.00 & 0.90 & 0.97 & 0.86 & 0.02 \\
Gemma-3-4B-it & 0.94 & 0.94 & 0.81 & 0.88 & 0.37 & 0.22 \\
Gemma-3-27B-it & 0.99 & 0.94 & 0.97 & 0.97 & 0.87 & 0.03 \\
\underline{GLM-4.7} & 1.00 & 1.00 & 0.93 & 0.97 & 0.82 & 0.02 \\
GPT-oss-120B & 1.00 & 1.00 & 0.91 & 0.84 & 0.90 & 0.00 \\
GPT-4o-mini & 1.00 & 0.99 & 0.90 & 0.96 & 0.90 & 0.02 \\
\underline{GPT-5-mini} & 1.00 & 0.99 & 0.92 & 0.96 & 0.93 & 0.03 \\
GPT-5 & 1.00 & 1.00 & 0.84 & 0.86 & 0.86 & 0.01 \\
Qwen3-4B & 0.97 & 0.96 & 0.92 & 0.97 & 0.86 & 0.05 \\
\underline{Qwen3-8B} & 0.97 & 0.96 & 0.91 & 0.97 & 0.88 & 0.02 \\
Qwen3-32B & 1.00 & 1.00 & 0.92 & 0.97 & 0.89 & 0.01 \\
Qwen3-235B-A22B & 1.00 & 1.00 & 0.92 & 0.98 & 0.91 & 0.02 \\
\hline
\end{tabular*}
\label{tab:judge_performance}
\end{table}

\subsubsection{Final Judge Selection}

Based on the analysis above, we selected three judges for the final evaluation setup: \textit{GPT-5-mini}, \textit{Qwen3-8B}, and \textit{GLM-4.7}. The selected judges are underlined in Table~\ref{tab:judge_performance}.

The selection was based on a combination of criteria:

\begin{itemize}
\item High F1 scores in distinguishing correct and incorrect samples.
\item Strong correlation with the consensus of all judges.
\item Low average bad scores on negative samples.
\item Diversification: selected models should represent different families to reduce the risk of systemic bias.
\end{itemize}

For the selected judges we computed the \textbf{Intraclass Correlation Coefficient}, $\mathrm{ICC}=0.96$. This shows that the selected judges give consistent scores, which allows them to be used as a reliable multi-judge evaluation system for subsequent experiments.

\section{Discussion}

Results presented in Table~\ref{tab:generation_performance} show that even small models are capable of producing results comparable to those of larger models. Although answer quality depends on model size, some sufficiently compact models still provide adequate generation quality for RAG systems. For our production system, we selected \textit{Qwen3-4B-Instruct-2507-Q5KM} due to its favourable trade-off between response quality and CPU inference latency.

\begin{table}[H]
\centering
\caption{Average evaluation metrics for different models. The best result is highlighted in bold, and the second-best result is underlined.}
\begin{tabular*}{\textwidth}{l@{\extracolsep{\fill}}cccc}
\hline
Model & Correct. & AnswerRelev. & Faithful. & Latency (s) \\
\hline
DeepSeek-R1-Distill-Qwen-7B-Q4KM & 0.40 & 0.54 & 0.58 & 297.4 \\
Llama-2-7B-Chat-Q4KM & 0.32 & 0.46 & 0.42 & 115.0 \\
Meno-tiny-1.5B-0.1-FP16 & 0.41 & 0.57 & 0.60 & \textbf{27.8} \\
Meno-lite-7B-Q4KM & 0.56 & 0.75 & 0.74 & \underline{31.4} \\
Mistral-7B-Instruct-v0.2-Q4KM & 0.47 & 0.61 & 0.63 & 115.5 \\
Phi-4-mini-Instruct-Q5KM & 0.48 & 0.68 & 0.67 & 44.2 \\
Qwen2.5-1.5B-Instruct-Q5KM & 0.46 & 0.61 & 0.63 & 49.1 \\
Qwen2.5-3B-Instruct-Q5KM & 0.54 & 0.73 & 0.72 & 31.8 \\
Qwen2.5-7B-Instruct-Q4KM & 0.64 & 0.85 & 0.77 & 66.5 \\
Qwen3-1.7B-Q4KM & 0.58 & 0.78 & 0.74 & 55.0 \\
Qwen3-4B-Instruct-2507-Q5KM & 0.71 & \textbf{0.89} & 0.80 & 70.9 \\
Qwen3-4B-Q5KM & 0.69 & 0.85 & 0.81 & 205.1 \\
Qwen3-8B-Q4KM & \underline{0.72} & 0.87 & \underline{0.83} & 339.3 \\
QVikhr-3-4B-Instruction-Q5KM & 0.59 & 0.64 & 0.71 & 254.4 \\
Saiga-Llama3-8B-Q4K & 0.60 & 0.79 & 0.72 & 92.2 \\
Saiga-Mistral-7B-Q4K & 0.44 & 0.53 & 0.54 & 257.1 \\
Vikhr-Llama-3.2-1B-Q5KM & 0.42 & 0.52 & 0.58 & 34.3 \\
\hline
GPT-5-mini & \textbf{0.73} & \underline{0.88} & \textbf{0.89} & -- \\
GPT-5-mini, No context & 0.47 & 0.86 & -- & --\\
\hline
\end{tabular*}
\label{tab:generation_performance}
\end{table}

From the results produced by the baseline model \textit{GPT-5-mini}, it becomes evident that the presence of contextual information significantly influences the accuracy of generated answers. This observation underscores a crucial characteristic of the dataset: generating correct responses requires reliance on external context rather than merely leveraging inherent model knowledge.

Since the dataset consists of queries written in Russian, we also analysed the language of the generated responses. Importantly, the models were not explicitly instructed to answer in Russian; only the input queries were provided in Russian. This setup allows us to observe the language preference of the models during generation.

\begin{figure}[ht]
\centering
\includegraphics[width=1.0\linewidth]{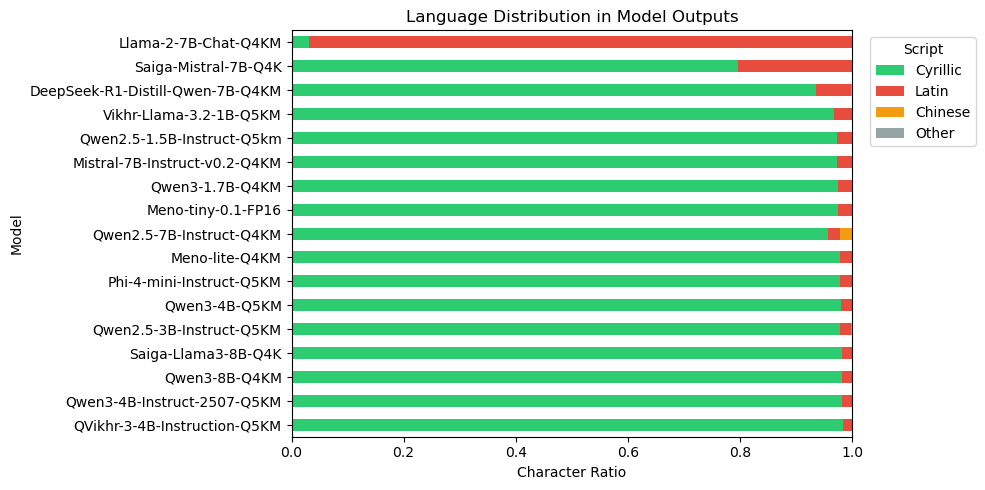}
\caption{Distribution of response languages across evaluated models.}
\label{fig:language_distribution}
\end{figure}

Figure~\ref{fig:language_distribution} shows the distribution of response languages across the evaluated models. The analysis highlights which models consistently produce answers in Russian and which tend to switch to English or generate mixed-language outputs.

To estimate the generation time, the models were run on a reduced subset of 50 samples. The evaluation was performed in a CPU-only environment without GPU acceleration to approximate typical local deployment conditions. The average generation time per answer for each model is presented in Table~\ref{tab:generation_performance}. The results demonstrate noticeable differences in inference speed across models. These measurements provide additional insights that can support model selection in practical deployment scenarios.

\section{Limitations}

This study provides extensive research and evaluation of SLMs specifically focusing on their performance as generative models. While providing valuable insights, several notable limitations constrain its scope:

\begin{enumerate}
    \item \textbf{Evaluation Focus}: The investigation focuses exclusively on SLMs' ability to generate text, disregarding their importance in determining optimal embedding techniques and ranking strategies within RAG systems.
    \item \textbf{Prompt Standardisation}: A uniform prompt was applied across all models and configurations, despite established knowledge that a single prompt may not suit every architecture effectively. Therefore, customising prompts for individual models could result in improved performance.
    \item \textbf{Task Restrictions}: Experimentation centres solely on RAG-oriented tasks, omitting broader applications beyond text generation. Incorporating diverse tasks would provide greater clarity on SLM applications.
    \item \textbf{Language Bias}: Results reflect performance in the Russian language alone, making generalisation to other languages uncertain without additional validation efforts.
\end{enumerate}

Addressing these shortcomings in subsequent studies promises to offer deeper insights into SLMs' true capabilities.

\section{Conclusion}

Although LLMs currently dominate the research landscape, SLMs continue to demonstrate considerable relevance across multiple domains while receiving considerably less attention. This study aims to bridge this gap by examining SLMs as generators for Russian-language RAG systems.

We introduce a curated benchmark compiled from diverse source datasets, ensuring a broad spectrum of topics to facilitate equitable comparisons among SLMs.

Subsequently, we systematically evaluated SLMs against this dataset, discovering that selected SLMs surpassed LLMs in terms of output quality while requiring substantially fewer computational resources. These findings suggest promising prospects for incorporating SLMs into practical applications.

We provide an extensive analysis of the SLMs' performance to show their advantages and trade-offs between quality and latency.

For future research, we recommend exploring improvements in embedding and reranking methodologies, along with expanding the scope of SLMs to address other tasks.



\bibliographystyle{splncs04}
\bibliography{mybibliography}

\end{document}